\documentclass[10pt]{article} 
\usepackage[preprint]{tmlr}

\usepackage{amsmath,amsfonts,bm}

\def\eqref#1{equation~\ref{#1}}

\def\1{\bm{1}}

\DeclareMathAlphabet{\mathsfit}{\encodingdefault}{\sfdefault}{m}{sl}
\SetMathAlphabet{\mathsfit}{bold}{\encodingdefault}{\sfdefault}{bx}{n}

\usepackage{hyperref}
\usepackage{url}
\usepackage{algorithm}
\usepackage{algorithmicx}
\usepackage{algpseudocode}
\usepackage{graphicx}
\usepackage{xcolor}
\usepackage{amssymb}
\usepackage{natbib}

\pagecolor{white} 
\title{Probabilistic Conformal Coverage Guarantees in Small-Data Settings}

\author{\name Petrus H. Zwart \email PHZwart@lbl.gov \\
      \addr Berkeley Synchrotron Infrared Structural Biology Program, Center for Advanced Mathematics in Energy Research Applications \& Molecular Biophysics and Integrated Bioimaging Division, 1 Cyclotron Road, Berkeley, CA 94720, USA
      }

\def\month{10}  
\def\year{2025} 
\def\openreview{\url{https://openreview.net/forum?id=XXXX}} 

\begin{document}

\maketitle

\begin{abstract}
Conformal prediction provides distribution-free prediction sets with guaranteed marginal coverage. However, in split conformal prediction this guarantee is training-conditional only in expectation: across many calibration draws, the average coverage equals the nominal level, but the realized coverage for a single calibration set may vary substantially. This variance undermines effective risk control in practical applications. Here we introduce the Small Sample Beta Correction (SSBC), a plug-and-play adjustment to the conformal significance level that leverages the exact finite-sample distribution of conformal coverage to provide probabilistic guarantees, ensuring that with user-defined probability over the calibration draw, the deployed predictor achieves at least the desired coverage.
\end{abstract}

\section{Introduction}

Modern machine learning applications increasingly demand rigorous uncertainty quantification for high-stakes decision-making. In domains such as medicine and autonomous systems, predictive models must not only be accurate but also know when they might be wrong—and quantify that uncertainty. Conformal prediction, introduced by \citet{vovk2005algorithmic}, has emerged as a powerful distribution-free framework to meet this need, offering finite-sample guaranteed coverage without strong model assumptions. In essence, given a desired miscoverage rate $\alpha$, conformal methods produce prediction sets (or intervals) that cover the true outcome with probability at least $1-\alpha$ \emph{marginally} over the randomness in the data.  That is, the guarantee holds when averaged over many independent draws of the data and test points: across repeated instances of the data-generating process, the coverage converges in the mean to the target level $1-\alpha$. This guarantee holds exactly in finite samples under exchangeability, setting conformal prediction apart from traditional parametric intervals and even Bayesian credible intervals \citep{AngelopoulosBates2023Gentle}.

\begin{figure}[!h]
    \centering
    \includegraphics[width=0.45\textwidth]{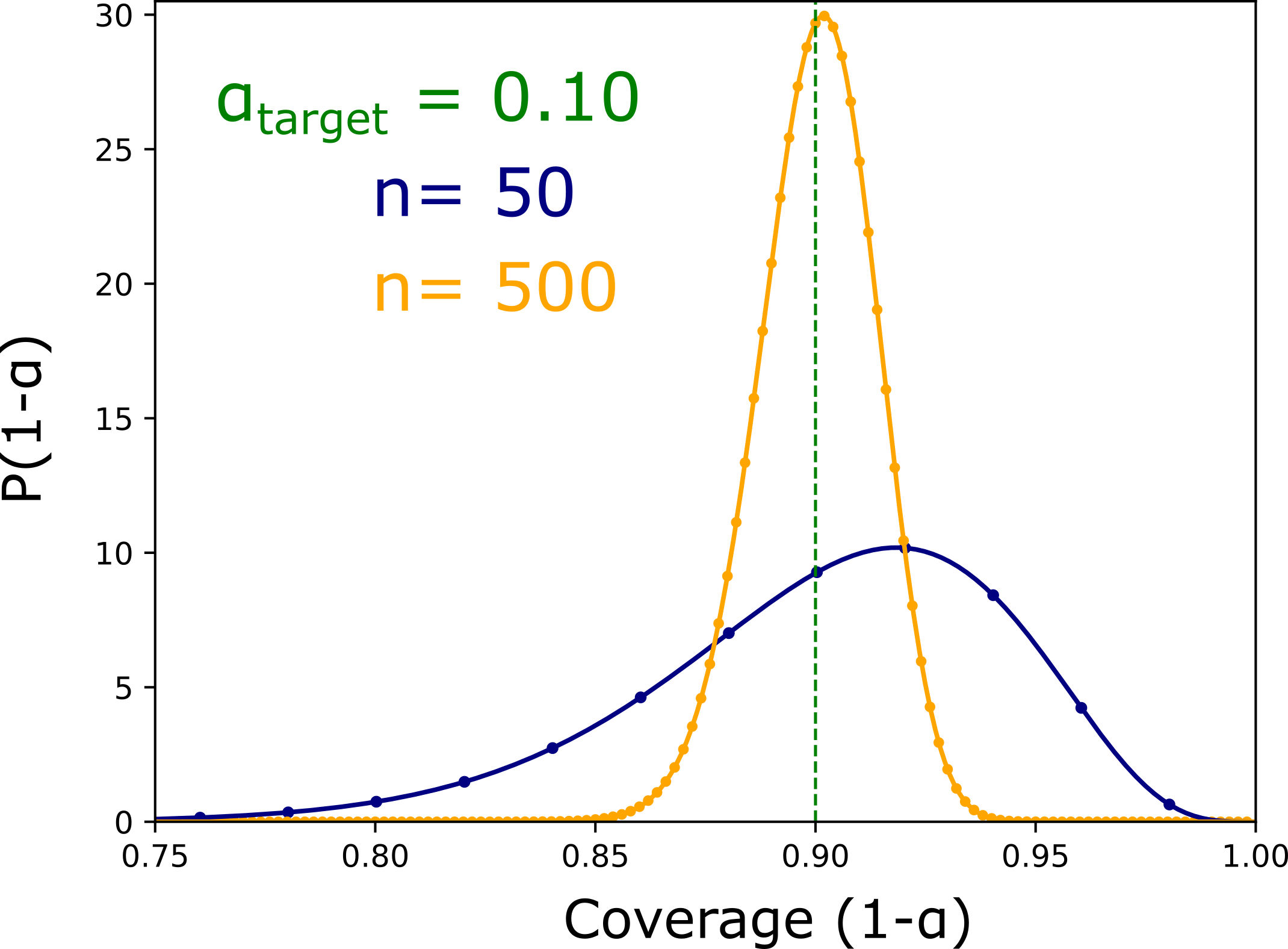}
    \caption{
    Distribution of empirical coverage $(1-\alpha)$ for conformal prediction with calibration sets of size $n=50$ (blue) and $n=500$ (orange). The vertical green dashed lines indicate the target coverage levels of $\alpha=0.10$. Despite using a substantially larger calibration set (n=500), considerable variability in coverage persists, with the distribution remaining notably wide around the target values. The finite-sample effects demonstrate that even with moderate calibration set sizes, coverage guarantees exhibit non-negligible uncertainty, highlighting the inherent variability in conformal prediction performance across different random splits of the data.
    }
    \label{fig:coverage_distribution}
\end{figure}

Among conformal approaches, split (inductive) conformal prediction has become popular for its computational efficiency as it requires a single training phase and uses a hold-out calibration set to size the prediction band. Split conformal inherits the formal guarantee that the \emph{expected} coverage equals the nominal level $1-\alpha$ across random splits of the calibration set:
\begin{equation}
\mathbb{E}_{\text{cal}}\!\big[\mathrm{coverage}\big] \ge 1-\alpha.
\end{equation}
However, this guarantee is marginal and for any \emph{particular} calibration set of size $n$, the realized coverage can deviate substantially from $1-\alpha$. In small-sample regimes, the distribution of empirical coverage is broad; some splits land far below the target even though the method is valid on average. Figure~\ref{fig:coverage_distribution} illustrates this variability; e.g., with $n=50$, substantial under-coverage is common, and the spread shrinks only gradually as $n$ grows. Recent theory makes this precise: in the infinite-test limit, the marginal coverage of split conformal follows a $\mathrm{Beta}$ law determined by $n$ and $\alpha$, and for finite test size it follows a Beta–Binomial law \citep{Marques2025Beta}. These results formalize that small-$n$ coverage is inherently variable, undermining operational trust when a single calibration split is deployed.

This limitation has spurred interest in guarantees that go beyond the simple marginal promise. In particular, \emph{Probably Approximately Correct} (PAC) coverage introduces a confidence parameter $\delta$ and mandates that with probability at least $1-\delta$ over any random calibration draw, the achieved coverage is near the nominal level \citep{Vovk2012PAC}:
\begin{equation}
\Pr\!\big(\text{coverage} \;\ge\; 1-\alpha-\varepsilon\big) \;\ge\; 1-\delta.
\end{equation}
\citet{Bian2023TrainingConditional} and others show that split conformal can admit such PAC guarantees without algorithmic changes. In practice, we seek these PAC guarantees rather than purely marginal ones, since decisions based on risk metrics require confidence in the performance on the present dataset, not only on average across repeated samples. 

Several strategies target obtaining PAC-style reliability guarantees. A classical route uses concentration inequalities to shrink the target level $\alpha$ by a tolerance $\epsilon$ so that
\begin{equation}
\Pr\!\big(\text{coverage} \;\ge\; 1-\alpha_{\text{target}}\big) \;\ge\; 1-\delta,
\end{equation}
implemented by setting $\alpha_{\text{adj}}=\alpha_{\text{target}}-\epsilon$. DKWM \citep{Massart1990DKWM} and Hoeffding \citep{Hoeffding1963Inequality} bounds yield $\epsilon = O\!\big(\sqrt{\ln(1/\delta)/n}\big)$, implying $O(\alpha^{-2})$ sample complexity when $\alpha$ is small. In practice, hitting $\alpha=0.01$ with high confidence can require thousands of calibration samples. This is often infeasible due to the cost of data acquisition. \citet{Yang2025Selection} demonstrate how such bounds can be used for selection/aggregation in conformal pipelines, but their looseness tends to produce over-wide sets in small-$n$ regimes. Alternative directions leverage Bayesian structure to reduce conservativeness \citep{Bersson2024SmallArea}, or seek relaxed forms of conditional validity \citep{Feldman2021OrthogonalQR, Chernozhukov2021Distributional, Gibbs2025Conditional, Fannjiang2022FeedbackCovariateShift}. In Mondrian-style conformal \citep{vovk2005algorithmic}, an additional complication arises: conditioning within strata (e.g., positives) makes the per-stratum calibration and inference counts random because class prevalence is unknown and only estimated from finite samples; this extra binomial variability amplifies finite-sample uncertainty in coverage guarantees.

In this communication, we address the challenge of finite-sample coverage variability by directly exploiting the exact distributional form of conformal coverage and construct PAC guarantees directly by operating on the coverage distribution. If $n$ calibration points are available and the nominal miscoverage is $\alpha$, the infinite-test coverage behaves as
\begin{equation}
C \;\sim\; \mathrm{Beta}\!\big(k,\, n+1-k\big), 
\quad k \;=\; \big\lceil (1-\alpha)(n+1)\big\rceil,
\end{equation}
with a Beta–Binomial analogue for finite test windows \citep{Marques2025Beta}. 
Building on this, we introduce the \emph{Small Sample Beta Correction} (SSBC), a plug-and-play adjustment that chooses
\begin{equation}
\alpha_{\mathrm{adj}}
\;=\;
\arg\max_{\alpha' \in \{k/(n+1)\} \,:\, \alpha' < \alpha_{\text{target}}}
\;\Big\{ \Pr\!\big(C(\alpha') \;\ge\; 1-\alpha_{\text{target}}\big) \;\ge\; 1-\delta \Big\}.
\end{equation}
That is, \(\alpha_{\mathrm{adj}}\) is the largest adjusted miscoverage rate such that, with probability at least \(1-\delta\) over \emph{any} calibration draw, the realized coverage meets or exceeds the target level. Because split-conformal quantiles lie on the discrete grid $\{k/(n+1)\}$, SSBC searches this grid; in the small-$n$ regime of interest, this is trivial to compute. Compared with concentration-based adjustments, SSBC yields markedly sharper guarantees—improving sample complexity from $O(\alpha^{-2})$ to $O(1/\alpha)$ up to logarithmic factors—while remaining model-agnostic and dead simple to implement. We also provide a feasibility analysis that flags $(\alpha_{\text{target}},\delta)$ pairs that are unattainable for a given $n$. Empirically, across synthetic and real data, including cases with small calibration sets ($n\approx 50$), SSBC tightens coverage control, converting \emph{coverage in expectation} into an \emph{operational guarantee} suitable for deployment when better control of assumed risk is required.

\section{Methods}
\subsection{Distribution of Coverage}\label{sec:distribution-of-coverage}
Following \citet{Marques2025Beta}, the distribution of conformal coverage can be characterized explicitly using Beta and Beta–Binomial laws, depending on the inference regime. When the number of test points approaches infinity, coverage randomness arises solely from calibration uncertainty. Let
\begin{equation}
    k \;=\; \left\lceil (1-\alpha)\,(n+1)\right\rceil ,
\end{equation}
where $n$ is the calibration set size. The limiting distribution of coverage $C$ is then
\begin{equation}
    C_\infty \;\sim\; \mathrm{Beta}\!\big(k,\, n+1-k\big).
\end{equation}
This arises because the empirical quantile induces a random acceptance threshold distributed according to a Beta law. With $m$ test points, coverage includes both calibration and finite-sample randomness:
\begin{equation}\label{eq:beta-binomial-vanilla}
    C_m \;=\; \frac{X}{m}, \qquad
    X \;\sim\; \mathrm{Beta\text{-}Binomial}\!\big(m;\, k,\, n+1-k\big),
\end{equation}
capturing calibration randomness through the Beta law and test-sample randomness through Binomial sampling.

\subsection{PAC Guarantees via the Small Sample Beta Correction}
We seek guarantees ensuring that, with high probability over calibration randomness, the realized coverage meets the target level:
\begin{equation}
    \Pr\!\left( C(\alpha_{\text{adj}}) \;\ge\; 1 - \alpha_{\text{target}} \right) \;\ge\; 1 - \delta .
\end{equation}
Because conformal quantile thresholds are defined by order statistics, the feasible adjusted levels lie on a discrete grid:
\begin{equation}\label{eq:quantile_grid}
    \alpha_{\text{adj}} \;\in\; \left\{ \frac{u}{\,n+1} : u = 1, \dots, n \right\},
\end{equation}
reflecting the granularity imposed by the calibration size $n$. SSBC selects the largest grid point satisfying the PAC constraint, i.e.,
\begin{equation}
    \alpha_{\mathrm{adj}}
    \;=\;
    \arg\max_{\substack{\alpha' \in \{\,u/(n+1):\,u=1,\dots,n\,\}\\ \alpha' < \alpha_{\text{target}}}}
    \Big\{ \Pr\!\big( C(\alpha') \;\ge\; 1 - \alpha_{\text{target}} \big) \;\ge\; 1 - \delta \Big\}.
\end{equation}
Given the appropriate distribution of coverage for the regime at hand (Beta for $C_\infty$ or Beta–Binomial for $C_m$), SSBC evaluates each grid candidate $\alpha' = u/(n+1)$ via
\begin{equation}
    p(u) \;=\; \Pr\!\left( C(\alpha') \;\ge\; 1 - \alpha_{\text{target}} \right),
\end{equation}
and returns the largest $\alpha_{\text{adj}} = u^\ast/(n+1)$ with $p(u^\ast) \ge 1 - \delta$. In this way, SSBC chooses the least conservative correction that still satisfies the desired PAC guarantee.

These results apply in a Mondrian (per-class) setting as well by replacing $n$ with the class-$j$ calibration size $n_j$. The training-conditional PAC guarantee is unchanged: with probability at least $1-\delta$ over the calibration draw within class $j$, the realized coverage meets $1-\alpha_{\text{target}}$. Operational questions about windowed error budgets and uncertain class mix can be handled, if desired, via a finite-sample evaluation that combines the within-class Beta–Binomial predictive with a prevalence confidence–predictive.

\begin{algorithm}[H]
\caption{Small Sample Beta Correction (SSBC)}
\begin{algorithmic}
\Require Target miscoverage $\alpha_{\text{target}} \in (0,1)$, calibration size $n \in \mathbb{N}$, confidence level $1-\delta$, regime $\in\{\infty,\ m\}$ (with window size $m$ if finite)
\State $\alpha_{\text{cal}} \gets 0$
\State $t \gets 1 - \alpha_{\text{target}}$
\For{$u = 1, \ldots, n$} \Comment{grid index; $\alpha' = u/(n{+}1)$}
    \State $\alpha' \gets \tfrac{u}{n+1}$
    \State $a \gets \left\lceil (1-\alpha')(n+1) \right\rceil \;=\; n+1-u$, \quad $b \gets n+1-a \;=\; u$
    \If{regime $= \infty$} \Comment{infinite-test coverage}
        \State $p_{\mathrm{tail}} \gets \Pr\!\big[ Z \ge t \big],\ \ Z \sim \mathrm{Beta}(a,b)$
    \Else \Comment{finite window of size $m$}
        \State $x^\star \gets \left\lceil t\,m \right\rceil$
        \State $p_{\mathrm{tail}} \gets \Pr\!\big[ X \ge x^\star \big],\ \ X \sim \mathrm{Beta\text{-}Binomial}(m;\,a,b)$
    \EndIf
    \If{$p_{\mathrm{tail}} \ge 1-\delta$}
        \State $\alpha_{\text{cal}} \gets \max(\alpha_{\text{cal}}, \alpha')$
    \EndIf
\EndFor
\If{$\alpha_{\text{cal}} = 0$}
    \State \Return \textsc{Infeasible}
\Else
    \State \Return $\alpha_{\text{cal}}$
\EndIf
\end{algorithmic}\label{algo:ssbc}
\end{algorithm}

\subsection{Feasibility \& Requirements}
When working with conformal prediction, it is valuable to understand the limits of what guarantees are mathematically achievable under a fixed calibration size $n$ and risk tolerance $\delta$. In this section we derive the minimal target miscoverage that can be certified, providing insight into the attainable range of guarantees. This analysis does not prescribe practical settings directly, but rather clarifies the boundary of what the theory allows, against which SSBC can then be judged.

To find the maximal supported coverage (equivalently, the minimal feasible \(\alpha_{\text{target}}\) at confidence \(1-\delta\)), we solve the boundary case where the probabilistic constraint is tight. The most conservative possible solution occurs at \(k=1\) (so \(\alpha_{\text{adj}}=\frac{1}{n+1}\); see \eqref{eq:quantile_grid}). This induces coverage distributions that depend on the inference window size \(m\), yielding a closed-form threshold for the infinite-window case and a useful approximation for finite \(m\).

With \(k=1\), \(C_\infty \sim \mathrm{Beta}(n,1)\) and
\begin{equation}
\Pr\!\big(C_\infty \ge 1-\alpha\big) \;=\; 1 - (1-\alpha)^{n}.
\end{equation}
Setting this to $1-\delta$ gives the exact feasibility threshold
\begin{equation}\label{eq:alpha_star_infty}
\alpha^\star_\infty(n,\delta) \;=\; 1 - \delta^{1/n}.
\end{equation}
This expression makes explicit the fundamental tradeoff: with a calibration set of size $n$, one cannot demand simultaneously vanishing miscoverage $\alpha$ and vanishing risk $\delta$. The threshold $\alpha^\star_\infty(n,\delta)$ represents the mathematical limit of what coverage and confidence can be jointly guaranteed.

For a finite window of size \(m\), the boundary still occurs at \(u=1\), so \(C_m \sim \mathrm{Beta\text{-}Binomial}(m,n,1)\). While there is no simple closed form for the boundary, it can be found numerically, and a practical first-order approximation (Appendix~A) is
\begin{equation}\label{eq:alpha_star_m}
\alpha^\star_m(n,\delta) \;\approx\; 1 - \delta^{1/n} \;+\; \sqrt{\frac{\delta^{1/n}\,\big(1-\delta^{1/n}\big)}{2\pi\,m}}.
\end{equation}
This \(m^{-1/2}\) correction captures the additional slack available for finite inference windows, and vanishes as \(m\to\infty\), recovering \eqref{eq:alpha_star_infty}.

The above relations yield practical insights in the type of confidence guarantees that are available for a given calibration set size, as the continuous threshold \(\alpha^\star_\infty\) is implementable only if it lies on or above the calibration grid:
\begin{equation}
\alpha^\star_\infty \;\ge\; \frac{1}{n+1}
\;\;\Longleftrightarrow\;\;
\delta \;\le\; \Big(1-\frac{1}{n+1}\Big)^{n} \;=\; \Big(\frac{n}{n+1}\Big)^{n}.
\end{equation}

\begin{figure}[htbp]
\centering
\includegraphics[width=0.5\textwidth]{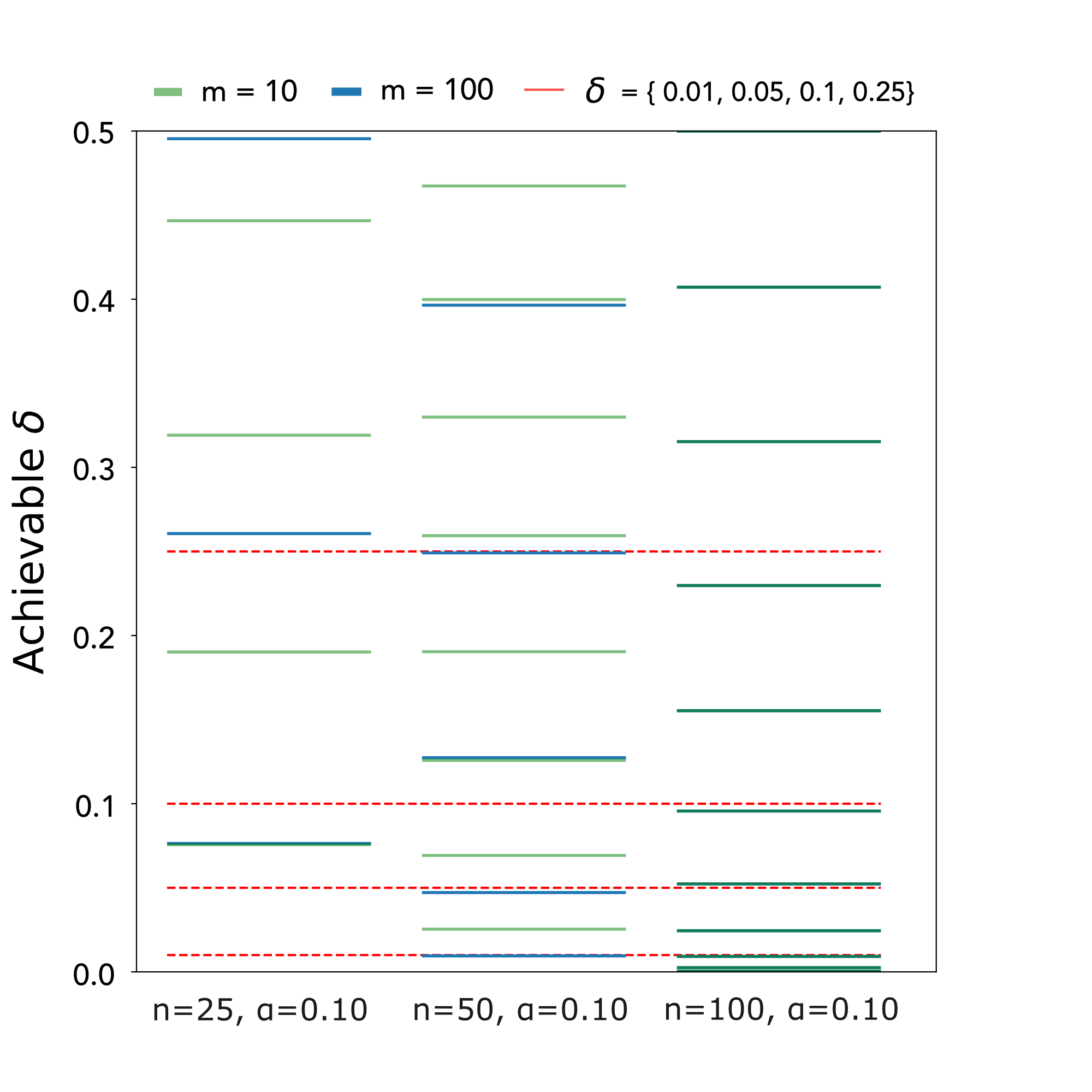}
\caption{Achievable confidence levels (\(\delta\)) under SSBC given calibration constraints. We show discrete rungs of attainable \(\delta\) for varying calibration sizes \(n\) at a fixed target miscoverage \(\alpha=0.10\), comparing finite inference windows \(m=10\) (green) and \(m=100\) (blue). Red dashed lines mark common reference levels \(\delta\in\{0.01,0.05,0.10,0.25\}\). The discrete structure reflects the grid constraint \(\alpha_{\text{adj}}\in\{u/(n{+}1)\}\); increasing \(n\) yields finer rungs, while larger \(m\) reduces the finite-window penalty.}
\label{fig:achievable_delta}
\end{figure}

Both \eqref{eq:alpha_star_infty} and \eqref{eq:alpha_star_m} describe \emph{continuous} feasibility thresholds; in practice, achievable levels are constrained to the discrete grid \(\alpha_{\text{adj}}=k/(n+1)\), where each \(k\) corresponds to a particular attainable \(\delta\), producing discrete \emph{rungs} rather than a continuum, Figure \ref{fig:achievable_delta}. Larger calibration sets \(n\) refine the grid (more rungs), and larger inference windows \(m\) reduce the finite-window slack, enabling more aggressive targets.

These relations summarize the trade-offs among calibration size \(n\), inference window \(m\), risk tolerance \(\delta\), and achievable coverage, and they provide simple rules for balancing statistical rigor against resource constraints in experimental design.

\section{Results}

\subsection{Numerical Simulations}
\label{subsec:simulations}
\begin{figure}[htbp]
  \centering
  \includegraphics[width=0.95\linewidth]{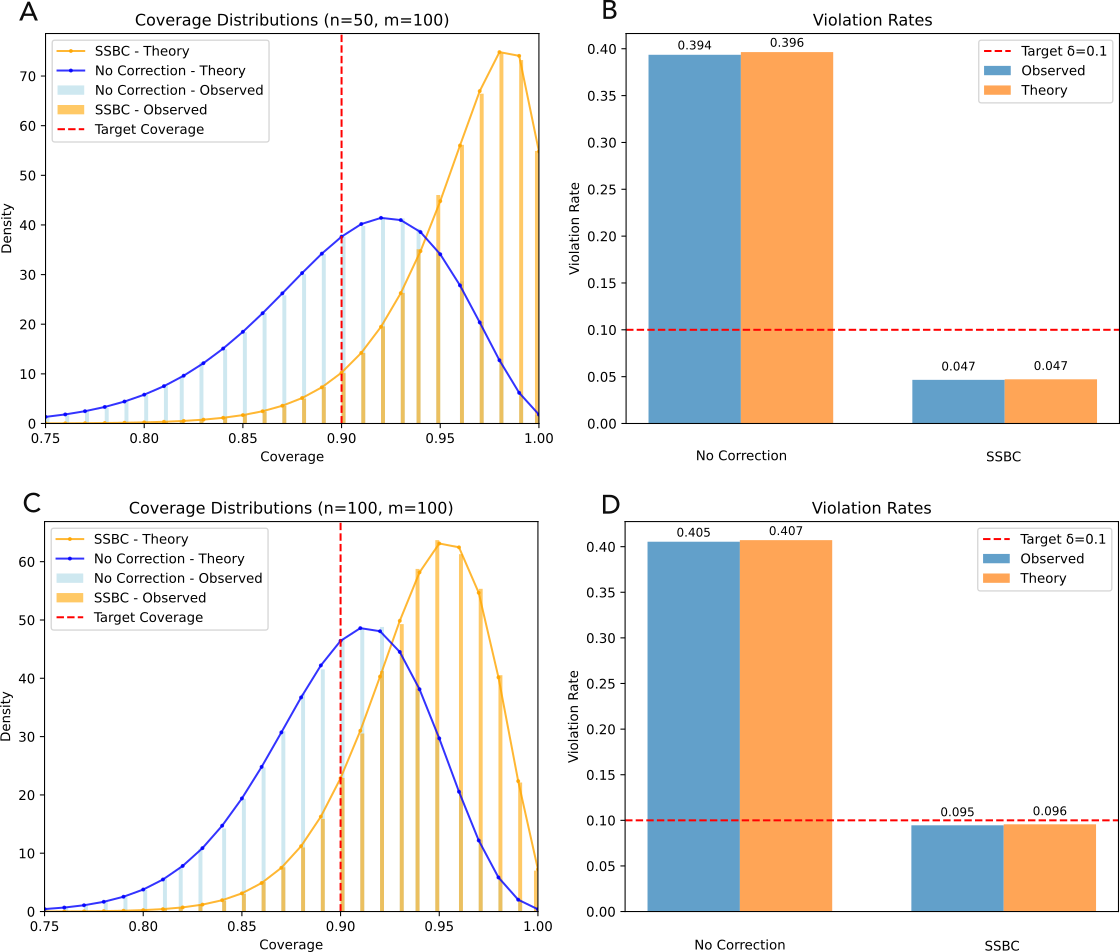}
  \caption{Monte Carlo validation of Small Sample Beta Correction (SSBC).
  Panels A–B use calibration size $n=50$ and inference window $m=100$; C–D use $n=100$, $m=100$.
  In all experiments the target miscoverage is $\alpha_{\text{target}}=0.1$ (vertical red line at $0.9$ coverage) and risk tolerance is $\delta=0.1$ (red dashed line in B,D).
  A,C: Coverage distributions under no correction (blue) and SSBC (orange). Solid lines show the theoretical Beta–Binomial (finite $m$) curves implied by the calibration quantile; shaded histograms are Monte Carlo estimates from 1 million runs with heavy-tailed non-conformity scores (absolute Cauchy variates $t_{\nu=1}$).
  B,D: Violation rates $\mathrm{Pr}(C<1-\alpha_{\text{target}})$ for each method: observed (blue) and theory (orange). 
  Without correction the violation rate is large ($\approx 0.394$–$0.407$). SSBC reduces violations to the design target, yielding $\approx 0.047$ for $n=50$ and $\approx 0.095$–$0.096$ for $n=100$, closely matching theory.}
  \label{fig:ssbc-sims}
\end{figure}

We empirically validate the Small Sample Beta Correction (SSBC) with Monte Carlo experiments that mirror the finite-sample distributional analysis used to derive the method. Each run draws $n$ calibration non-conformity scores as an absolute Cauchy random variable $|t_{\nu=1}|$, and computes the calibration quantile twice --- once at the nominal level $\alpha_{\text{target}}$ and once at the SSBC-adjusted level $\alpha_{\text{adj}}\in\{u/(n+1)\}_{k=1}^n$ chosen to satisfy $\mathrm{P}(C\ge 1-\alpha_{\text{target}})\ge 1-\delta$, and then estimates coverage on an inference window of size $m$ using fresh draws. The theoretical curves in Fig.~\ref{fig:ssbc-sims} come from the Beta–Binomial law for finite $m$ (Beta in the $m\to\infty$ limit) with parameters induced by the order statistic of the calibration threshold.

We report two configurations: $(n,m)=(50,100)$ and $(100,100)$ with $\alpha_{\text{target}}=0.1$ and $\delta=0.1$ over 1 million runs. Without correction, coverage varies widely and the violation rate $\mathrm{P}(C<0.9)$ is large -- $\approx 0.394$–$0.407$ -- consistent with the Beta–Binomial theory for the unadjusted calibration quantile. 
SSBC shifts the coverage distribution to the right and concentrates its mass above the target, cutting violations to the design level: $\approx 0.047$ for $n=50$ and $\approx 0.095$–$0.096$ for $n=100$. The fact that this design misses the requested $\delta = 0.1$ level is due to the fact that the first feasible rung below $0.1$, is $0.047$, Figure \ref{fig:achievable_delta}.

Taken together, these results confirm that SSBC delivers the intended PAC-style guarantee at the requested $\delta$ while standard split conformal, although marginally valid in expectation, fails to control single-draw risk at small $n$. 

\subsection{Image Segmentation}
To demonstrate SSBC's effectiveness with small calibration sets, we applied the method to 3D cryo-electron tomography (cryoET) segmentation tasks. The dataset \citep{Zens2024ECM} consists of cryoET data from EMPIAR \citep{Iudin2023EMPIAR} containing rich cellular ultrastructure. Here, we focus on collagen fibers only for demonstration purposes, which were segmented using minimally annotated training data. The segmentation was performed using an ensemble of 5 3D-SMSNets from the DLSIA \& qlty packages \citep{Roberts:yr5117,ZWART2024100696}. The class-conditional calibration set contained 4337 pixels, and was partitioned in smaller sets to highlight coverage variability in extremely small calibration sets, at reasonable coverage, $\alpha_{\mathrm{target}}=0.05$.

\begin{figure}[htbp]
    \centering
    \includegraphics[width=0.95\textwidth]{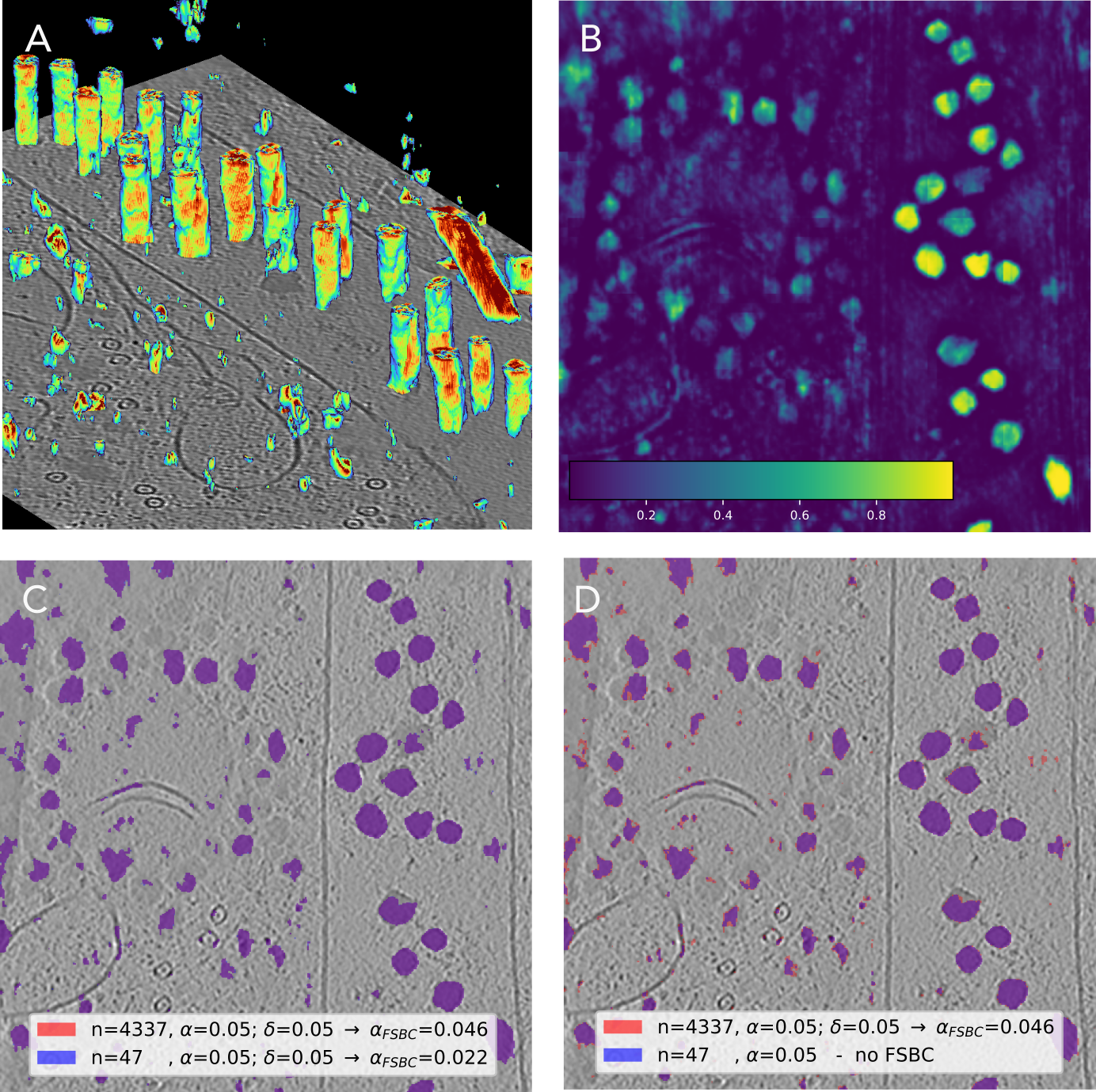}
    \caption{
        \textbf{Small Sample Beta Correction (SSBC) improves conformal prediction reliability with limited calibration data.} 
        A: 3D volumetric segmentation visualization with confidence-weighted prediction sets overlaid on microscopy data. Cylindrical structures show varying prediction confidence levels indicated by the color scale. 
        B: 2D probability map slice displaying prediction confidence for detected objects, with the color bar ranging from 0.2 (purple, low confidence) to 0.8+ (yellow, high confidence). 
        C: Conformal prediction sets using SSBC correction with calibration sets of n=4337 (red overlay, $\alpha_{SSBC}=0.046$) and n=47 (blue overlay, $\alpha_{SSBC}=0.022$) samples, both targeting $\alpha=0.05$ and $\delta=0.05$. Despite the 92$\times$ difference in calibration size, both provide reliable coverage of detected objects. 
        D: Standard conformal prediction without SSBC correction shows under-coverage for the small calibration set (n=47, blue), demonstrating the coverage variability problem that SSBC addresses. The SSBC method enables reliable uncertainty quantification even with severely limited calibration data, crucial for applications where labeled samples are expensive or scarce.
    }
    \label{fig:fsbc_demonstration}
\end{figure}

The segmentation model produces confidence scores for detected collagen fibers, which we visualize through 2D probability maps (Figure \ref{fig:fsbc_demonstration}B) and 3D confidence-weighted prediction sets overlaid on the cryoET data (Figure \ref{fig:fsbc_demonstration}A). The color scale ranges from low confidence (purple, 0.2) to high confidence (yellow, 0.8+). Figure 2C shows SSBC-corrected conformal prediction sets for both calibration sizes. Despite the dramatic difference in calibration data set size, SSBC enables both the large ($\alpha_\mathrm{SSBC} = 0.046$) and small ($\alpha_\mathrm{SSBC}  = 0.022$) calibration sets to yield virtually identical predicted label sets across the full image. The SSBC correction ensures that the severely limited calibration set maintains coverage guarantees.

In contrast, Figure \ref{fig:fsbc_demonstration}D demonstrates standard conformal prediction without SSBC correction applied to the same small calibration set (n=47), where the predicted labels without the correction is less permissive in what to include. This under-coverage for the small calibration set illustrates the coverage variability problem that SSBC addresses.

This application showcases SSBC's practical value in domains where labeled samples are expensive or scarce, such as structural biology where manual annotation of 3D cryoET data is extremely time-intensive. Rather than requiring thousands of manually annotated samples for reliable conformal prediction, SSBC enables robust uncertainty quantification even with severely limited calibration data from minimal annotations.

\subsection{Molecular Solubility}
Drug discovery relies heavily on accurate prediction of molecular properties, particularly aqueous solubility, which directly impacts bioavailability, formulation, and therapeutic efficacy. Poor solubility represents one of the primary causes of drug candidate failure, with estimates suggesting that up to 40\% of marketed drugs and 90\% of development pipeline compounds suffer from solubility limitations \citep{Kalepu2015Insoluble, Xie2024Solubilization}. Traditional experimental determination of solubility is time-intensive and expensive, often requiring weeks of synthesis and testing per compound, making computational prediction essential for efficient pharmaceutical development.
We applied SSBC to conformal quantile regression for molecular solubility prediction using a transformer-based architecture trained on SMILES molecular representations. The model pipeline (Figure 3A) converts molecular structures to embeddings, processes them through a transformer encoder, and outputs three quantile predictions (25th, 50th, and 75th percentiles) to capture prediction uncertainty. This quantile regression approach provides natural prediction intervals that, when combined with conformal corrections, deliver rigorous uncertainty quantification for solubility estimates that can guide downstream decision making what to synthesize.
We trained our model on the AqSolDB dataset, that contains 9,982 molecules with experimentally determined LogS values spanning the full range of pharmaceutical relevance \citep{Sorkun2019AqSolDB}. We partitioned this into 998 molecules for calibration purposes, 998 molecules for evaluation and the remaining ones for model training. The large calibration set enables robust assessment of conformal methods across various calibration sizes via subsampling this data, while maintaining the full evaluation set for consistent comparison.

A visual depiction of the calibration behavior for $n_{\mathrm{cal}} = 50$ is shown in Figure 3B. The distribution of quantile regression corrections demonstrates SSBC's ability to achieve reliable coverage, $\delta_{\mathrm{obs}} = 0.064$, where $\delta_{\text{obs}} := \Pr(C < 1-\alpha_{\text{target}})$ is the observed violation rate. This is achieved with 20-fold fewer calibration samples compared to the 998-sample baseline, while closely matching theoretical predictions. SSBC produces corrections concentrated between $-0.05$ and $0.3$, indicating modest adjustments to the original quantile predictions that maintain reasonable interval widths. In contrast, DKWM generates corrections spanning $0.0$ to $0.5$, with substantial density at higher values that produce unnecessarily wide prediction intervals. This distributional difference reflects SSBC's ability to make precise adjustments based on the exact finite-sample distribution of coverage, while DKWM's concentration-based bounds impose uniform over-corrections regardless of the calibration scenario. The targeted correction distribution from SSBC translates directly to more efficient prediction intervals that achieve the target reliability without sacrificing predictive precision.

DKWM provides overly conservative coverage estimates ($\delta_{\mathrm{obs}} = 0.016$), yielding intervals that are wider than operationally necessary. While excessive conservatism appears safe, this represents a decision that should be governed by domain-specific policy and risk tolerance rather than imposed by suboptimal theoretical approximations. The calibrated prediction intervals (Figure 3C) demonstrate proper coverage across the full solubility range encountered in drug discovery, spanning from highly insoluble compounds (LogS $< -8$) to readily soluble molecules (LogS $> -2$). The median prediction line closely tracks empirical quantiles, validating both model calibration and SSBC correction effectiveness.

\begin{figure}[htbp]
    \centering
    \includegraphics[width=\textwidth]{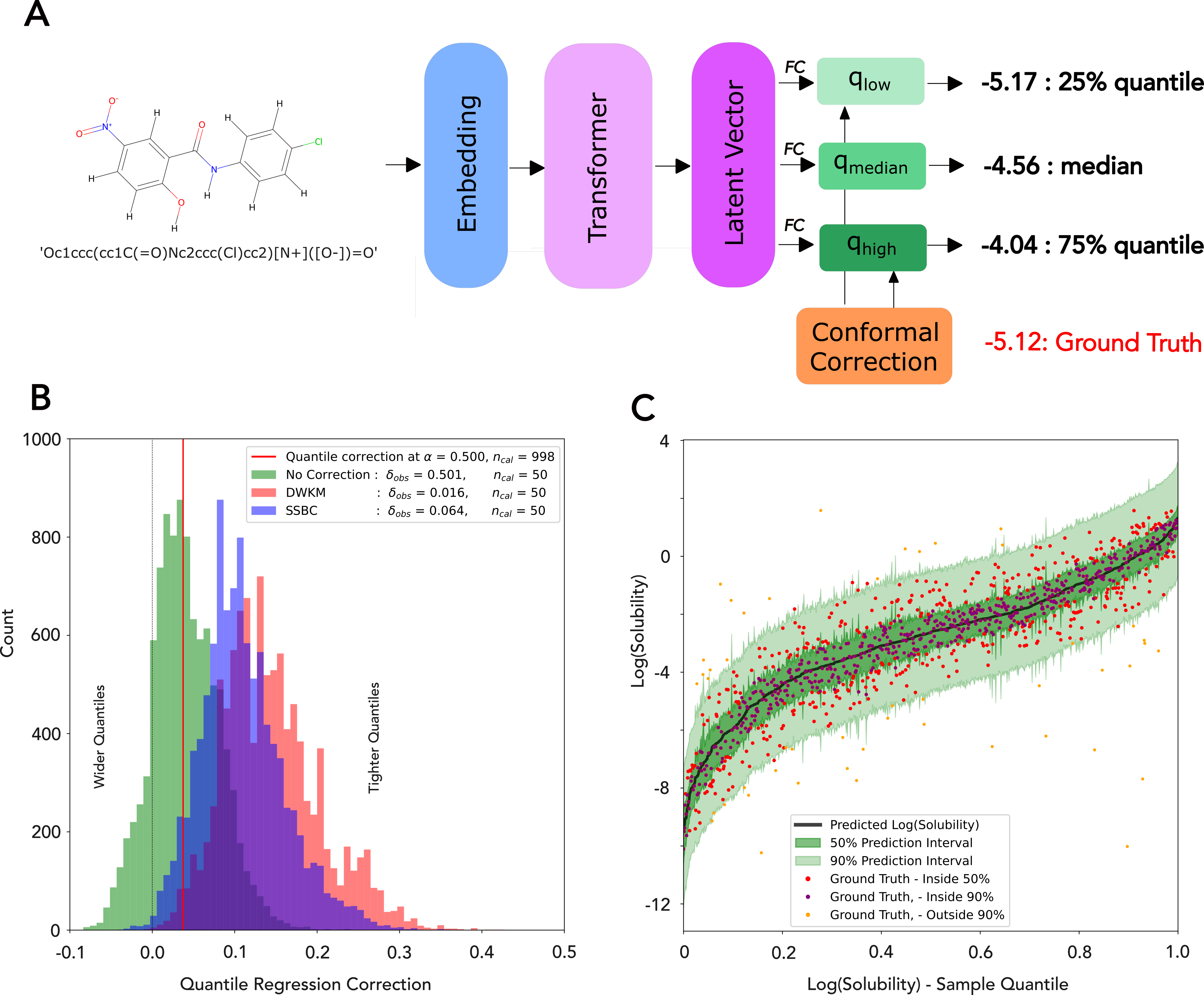}
    \caption{
        Conformalized quantile regression for molecular solubility prediction with efficient calibration. (A) Pipeline overview: SMILES strings are converted to molecular embeddings, processed through a transformer encoder, and mapped to a latent vector that feeds three prediction heads $(q_\mathrm{low}, q_\mathrm{median}, q_\mathrm{high}$ for 25th, 50th, and 75th percentiles. Conformal correction adjusts interval widths to achieve target coverage. Example shows chloramphenicol with predicted quantiles and ground truth (-5.12). (B) Distribution of quantile regression corrections across methods using 50 calibration samples compared to baseline with 998 samples. Red line indicates target 50\% coverage threshold in the full calibration dataset. SSBC achieves a lower-violation rate of the target $\alpha=0.5$ of $\delta_{\mathrm{obs}} = 0.064$ with 20-fold fewer calibration samples, while maintaining similar performance to the full dataset baseline. The requested miscoverage risk was specified as 0.1. The DKWM method is overly conservative ($\delta_{\mathrm{obs}} = 0.016$), producing wider intervals. (C) Calibrated prediction intervals on holdout data ($n = 995$ molecules). Green shading shows 50\% prediction intervals, light green shows 90\% intervals. Red and orange points indicate ground truth values within respective intervals, demonstrating proper coverage across the full solubility range. The median prediction line closely tracks the empirical quantiles, validating model calibration.}
    \label{fig:delta_demonstration}
\end{figure}

\begin{table}[h]
\centering
\begin{tabular}{c|c|cc|cc}
\hline
$n_\mathrm{cal}$ & None& DKWM-Obs & DKWM-Theory&  SSBC-Obs & SSBC-Theory \\
\hline
25 & 0.4979 & 0.0135 & 0.0164 & 0.0579 & 0.0539 \\
50 & 0.4939 & 0.0136 & 0.0207 & 0.0581 & 0.0460 \\
75 & 0.5041 & 0.0129 & 0.0176 & 0.0844 & 0.0827 \\
100 & 0.5016 & 0.0110 & 0.0177 & 0.0879 & 0.0816 \\
125 & 0.5060 & 0.0129 & 0.0204 & 0.0669 & 0.0761 \\
150 & 0.5029 & 0.0110 & 0.0190 & 0.0709 & 0.0825 \\
175 & 0.5000 & 0.0080 & 0.0200 & 0.0712 & 0.0867 \\
200 & 0.5078 & 0.0090 &0.0177 &  0.0691 & 0.0895 \\
\hline
\end{tabular}
\caption{Observed miscoverage rates when applying various corrections, based on 10k random calibration of indicated size. $\alpha_{\mathrm{target}}$ was 0.50 and requested $\delta$ was 0.10 --- due to discretization induced feasibility rungs and the possibility of slight covariate shift between calibration and test sets, some error is expected. Coverage violations when no correction is applied is equal to 50\%, as expected. The DKWM correction yields very conservative corrections, and observed values are in line with what is expected from the Beta distribution at the DKWM adjusted $\alpha$ and the desired miscoverage of 50\%. SSBC provides a less conservative correction, and observed miscoverage rates are in line with theory as well.}
\label{tab:your_label}
\end{table}

\section{Discussion}

The results presented here demonstrate that the Small Sample Beta Correction (SSBC) provides a practical, distribution-free and \emph{drop--in} mechanism for ensuring that practitioners obtain the actual coverage they require for their downstream, application, even in small calibration sample settings. The key contribution is a lightweight correction that transforms marginal guarantees into Probably Approximately Correct (PAC) guarantees, with user-tuneable risk. 

When comparing SSBC and concentration-based approaches such as Dvoretzky--Kiefer--Wolfowitz--Massart (DKWM) and Hoeffding-style corrections, one finds that the latter provide mathematically valid guarantees, but they do so by being deliberately over-conservative, masking the true risk profile behind inflated margins. As shown in both the simulations and molecular solubility experiments, DKWM-based corrections yield extremely wide intervals with violation probabilities far below the requested level. This ``safety by fiat'' places risk management decisions in the hands of concentration inequalities designed to minimize the risk across the full distribution, rather than having it controlled it at the quantile of interest by the practitioner. SSBC offers fine-grained control, where one can select their risk tolerance $\delta$, and the method ensures guarantees that match this operational requirement without imposing excessive conservatism.

Monte Carlo experiments validate this trade-off. Without correction, conformal coverage is highly variable, with observed violation rates around 40\% for target coverage of 90\%. SSBC shifts the distribution appropriately, aligning empirical violation rates with the specified $\delta$ (e.g., $\approx 0.047$ for $n = 50$, consistent with the quantile discretization grid). In contrast, DKWM corrections, while technically valid, produce violation rates an order of magnitude smaller than requested. These findings confirm that SSBC is not just mathematically principled but also computationally practical, requiring only modest calibration data to deliver meaningful guarantees.

The cryo-electron tomography (cryoET) segmentation experiments illustrate the impact of SSBC in high-cost, small-sample settings. With calibration sets as small as 47 pixels, SSBC yields label sets nearly indistinguishable from those calibrated with thousands of points, effectively neutralizing the instability of small calibration samples. Without this correction, conformal prediction systematically undercovers in this regime, potentially excluding biologically meaningful details.

Similarly, in molecular solubility prediction, SSBC enables robust quantile–regression intervals with \emph{20× fewer} calibration samples than the baseline. The correction maintains proper coverage across the solubility spectrum, whereas DKWM (Dvoretzky–Kiefer–Wolfowitz–Massart) defaults to excessive conservatism. These results show that SSBC generalizes across use cases -- classification to regression -- underscoring its versatility. In many PAC constructions, guarantees are obtained via DKWM-type concentration inequalities; by contrast, our approach works directly with the \emph{distribution of coverage}, which empirically yields tighter behavior.

\subsection{Conclusion}
SSBC provides a simple way to obtained probabilistically guaranteed coverage control by adjusting the effective calibration level \(\alpha\). It leverages the fact that, in split conformal prediction, the distribution of coverage is known, allowing practitioners to choose the calibration rung that achieves the desired probably approximately correct (PAC) guarantee for their \((n,m,\delta)\) without guesswork.

Unlike DKWM-style bounds, which trade efficiency for caution, SSBC delivers distribution-free coverage without relying on such conservative inequalities and returns risk control to users, who set \(\delta\) to match their domain’s tolerance. The method is lightweight, easy to implement, and effective across applications from structural imaging to molecular property prediction. In short, SSBC is a small but impactful adjustment that makes conformal prediction accountable where it matters most.

\bibliographystyle{plainnat}
\bibliography{ssbc}


\clearpage

\providecommand{\PR}{\Pr}
\providecommand{\Bin}{\mathrm{Bin}}
\providecommand{\Imap}{I_m}              
\providecommand{\Iinf}{I_\infty}         
\providecommand{\PhiN}{\Phi}             

\section*{Appendix: Laplace Approximation for Minimal Achievable Coverage with Finite Inference Windows}
\addcontentsline{toc}{section}{Appendix: Laplace Approximation for Finite Windows}

\noindent
We study the finite-$m$ correction to the minimal achievable target miscoverage $\alpha_m^\star$ for the Beta--Binomial case.
Define the Beta-mixture tail functional
\[
\Imap(\alpha) \;:=\; \int_{0}^{1} \PR\!\big(\Bin(m,p)\,\ge\, m(1-\alpha)\big)\; n\,p^{\,n-1}\,dp,
\]
i.e., the tail probability $\PR(\Bin(m,p)\ge m(1-\alpha))$ averaged under the $\mathrm{Beta}(n,1)$ density $n p^{n-1}$.
We determine $\alpha_m^\star$ from the target equation $\Imap(\alpha_m^\star)=1-\delta$ and derive its large-$m$ expansion.

\begin{equation}
\Imap(\alpha_m^\star) \;=\; 1 - \delta.
\label{eq:target}
\end{equation}

\begin{equation}
\Imap(\alpha)
\;=\;
\int_{0}^{1}
\PR\!\big(\Bin(m,p)\,\ge\, m(1-\alpha)\big)\; n\,p^{\,n-1}\,dp.
\end{equation}

\begin{equation}
\Iinf(\alpha)
\;:=\;
\lim_{m\to\infty}\Imap(\alpha)
\;=\;
\int_{1-\alpha}^{1} n\,p^{\,n-1}\,dp
\;=\;
1 - (1-\alpha)^{n}.
\end{equation}

\noindent
Let $\alpha_0 := 1-\delta^{1/n}$ and $x_0 := 1-\alpha_0=\delta^{1/n}$. We analyze $\Imap(\alpha)$ near the transition point
by introducing the rescaled coordinate
\begin{equation}
p \;=\; x_0 + \frac{y}{\sqrt{m}},
\qquad
y = O(1)\ \text{as } m\to\infty,
\qquad
x_0 \;=\; \delta^{1/n}.
\end{equation}

\noindent
A normal (Berry--Esseen) approximation to the binomial tail yields, uniformly for $y$ in compact sets,
\begin{equation}
\PR\!\big(\Bin(m,p) \ge m x_0\big)
\;=\;
\PhiN\!\left(
\frac{p - x_0}{\sqrt{p(1-p)/m}}
\right)
\;+\; O(m^{-1/2})
\;=\;
\PhiN\!\left(
\frac{y}{\sqrt{x_0(1-x_0)}}
\right)
\;+\; O(m^{-1/2}).
\end{equation}

\noindent
In the $O(m^{-1/2})$ transition layer, the Beta weight varies slowly:
\begin{equation}
n\,p^{\,n-1}
=
n\!\left(x_0 + \frac{y}{\sqrt{m}}\right)^{\!n-1}
=
n\,x_0^{\,n-1} \;+\; O(m^{-1/2}),
\qquad
dp \;=\; \frac{dy}{\sqrt{m}}.
\end{equation}

\noindent
Thus the contribution from the transition region can be written as
\begin{equation}
\Imap(\alpha)
\;=\;
\frac{n\,x_0^{\,n-1}}{\sqrt{m}}
\int_{-\infty}^{\infty}
\PhiN\!\left(\frac{y}{\sqrt{x_0(1-x_0)}}\right)\,dy
\;+\; O(m^{-1})
\quad\text{(local form near the step).}
\end{equation}

\noindent
Introduce the steepness parameter
\begin{equation}
\beta \;=\; \sqrt{\frac{m}{x_0(1-x_0)}}.
\end{equation}

\noindent
A standard smoothed-step (Watson/Laplace) expansion gives, as $\beta\to\infty$,
\begin{equation}
\int_{-\infty}^{\infty} \PhiN(\beta y)\, g(y)\,dy
\;=\;
\int_{0}^{\infty} g(y)\,dy
\;+\;
\frac{g(0)}{\beta\sqrt{2\pi}}
\;+\; O(\beta^{-2}),
\qquad
\text{for smooth } g.
\end{equation}

\noindent
Applying this with $g(y)\equiv n\,x_0^{\,n-1}$ and combining the outer regions, we obtain the first-order correction
to $\Imap(\alpha)$:
\begin{equation}
\Imap(\alpha)
\;=\;
1 - (1-\alpha)^{n}
\;+\;
\frac{n\,x_0^{\,n-1}}{\sqrt{2\pi m}}\,
\frac{1}{\sqrt{x_0(1-x_0)}}
\;+\;
O(m^{-1}).
\end{equation}

\noindent
To solve \eqref{eq:target}, write $\alpha_m^\star = \alpha_0 + \varepsilon$ and linearize the leading term:
\begin{equation}
1 - (1 - \alpha_m^\star)^{n}
\;=\;
1 - (1 - \alpha_0 - \varepsilon)^{n}
\;\approx\;
1 - (1-\alpha_0)^{n}
\;+\;
n(1-\alpha_0)^{n-1}\,\varepsilon.
\end{equation}

\noindent
Plugging this and $x_0=1-\alpha_0$ into the corrected expression at level $1-\delta$ gives
\begin{equation}
1 - \delta + n(1-\alpha_0)^{n-1}\,\varepsilon
\;=\;
1 - \delta
\;+\;
\frac{n\,x_0^{\,n-1}}{\sqrt{2\pi m}}\,
\frac{1}{\sqrt{x_0(1-x_0)}}.
\end{equation}

\noindent
Canceling the common factor $n(1-\alpha_0)^{n-1} = n x_0^{\,n-1}$ yields
\begin{equation}
n(1-\alpha_0)^{n-1}\,\varepsilon
\;=\;
\frac{n\,x_0^{\,n-1}}{\sqrt{2\pi m}}\,
\frac{1}{\sqrt{x_0(1-x_0)}}.
\end{equation}

\begin{equation}
\varepsilon
\;=\;
\sqrt{\frac{\alpha_0(1-\alpha_0)}{2\pi m}},
\qquad
\alpha_0 = 1 - \delta^{1/n}.
\end{equation}

\noindent
Therefore, the minimum achievable target miscoverage in a finite inference window of size $m$ is
\begin{equation}
\alpha_m^\star
\;=\;
1 - \delta^{1/n}
\;+\;
\sqrt{\frac{(1-\delta^{1/n})\,\delta^{1/n}}{2\pi m}}
\;+\;
O(m^{-1}).
\end{equation}

\noindent
The $O(m^{-1/2})$ correction is positive, reflecting that finite $m$ requires a slightly more conservative
(i.e., larger) $\alpha$ than the $m\to\infty$ limit $\alpha_0=1-\delta^{1/n}$; the correction vanishes as $m\to\infty$,
recovering the infinite-window result. Further validation of these expressions come from  computing the exact finite-window feasibility floor \(\alpha_m^\star\) by numerically solving the Beta–Binomial mixture equation and compare it to the asymptotic prediction.
Across interior settings, \(\alpha_m^\star-\alpha_0\) scales linearly in \(1/\sqrt{m}\) with fitted slope close to \(\sqrt{\alpha_0(1-\alpha_0)/(2\pi)}\), varying between 50\% and 150\% of the expected value. This variability is not a major surprise, given that exact values are influenced by calibration set size discretizations.

\section*{Appendix: PAC Guarantees in Inference Windows With Class Prevalence Uncertainty}
\label{subsec:prevalence-uncertainty}
In deployment the number of miscoverage count \emph{counts} per evaluation interval and per class can be important for operational purposes, not just per-item rates. Because the class mix varies from window to window, the number of class-$j$ items $m_j$ is random; controlling only per-item (Mondrian) coverage can still blow a window-level error budget in a busy window or be overly conservative in a quiet one. We therefore formulate a window-level, class-conditional target that accounts for calibration noise and class-mix uncertainty, and we compute the corresponding small-sample adjustment via Algorithm~\ref{algo:ssbc}.

We work with a single class $j$ \citep{vovk2005algorithmic}. Let $k$ be the training size with $k_j$ items of class $j$. On the calibration split of size $n$, let $n_j$ be the number of class-$j$ points and $s_j$ the number of class-$j$ miscoverages at the split-conformal cutoff,
\begin{equation}
s_j \;=\; n_j - \Big\lceil (1-\alpha)(n_j+1) \Big\rceil + 1,
\end{equation}
the smallest integer consistent with target miscoverage $\alpha$. In an evaluation window of size $m$, write $m_j$ for the (random) number of class-$j$ items and $e_j$ for their miscoverages.

From $(k_j,k)$ the Clopper--Pearson inversion \citep{clopper1934use} yields a confidence distribution $p_j\sim\mathrm{Beta}(k_j,\,k-k_j)$ for class-$j$ prevalence (exact frequentist coverage; numerically close to a uniform-prior Bayesian analysis \citep{brown2001interval,casella2002statistical}). Marginalizing $m_j\mid p_j\sim\mathrm{Binomial}(m,p_j)$ against this Beta law gives the confidence–predictive for the future class count:
\begin{equation}\label{eq:bb-mj-no-pseudo}
m_j \sim \mathrm{Beta\text{-}Binomial}\!\big(m;\,k_j,\,k-k_j\big),
\qquad
\Pr(m_j=r)=\binom{m}{r}\,\frac{B(r+k_j,\,m-r+k-k_j)}{B(k_j,\,k-k_j)},
\quad r=0,\dots,m,
\end{equation}
with deterministic boundaries when $k_j\in\{0,k\}$. Analogously, $(s_j,n_j)$ induces $\theta_j\sim\mathrm{Beta}(s_j,\,n_j-s_j)$ for the class-$j$ miscoverage rate; averaging $e_j\mid (m_j=r,\theta_j)\sim\mathrm{Binomial}(r,\theta_j)$ over this Beta law yields
\begin{equation}
\Pr(e_j=e \mid m_j=r)
=
\binom{r}{e}\,\frac{B(e+s_j,\,r-e+n_j-s_j)}{B(s_j,\,n_j-s_j)},
\qquad e=0,\dots,r.
\end{equation}
Combining the two pieces gives the joint predictive over $(e_j,m_j)$:
\begin{equation}\label{eq:joint-lc}
\Pr(e_j=e,\,m_j=r)
=
\binom{m}{r}\,\frac{B(r+k_j,\,m-r+k-k_j)}{B(k_j,\,k-k_j)}
\cdot
\binom{r}{e}\,\frac{B(e+s_j,\,r-e+n_j-s_j)}{B(s_j,\,n_j-s_j)},
\quad 0\le e\le r\le m.
\end{equation}

For a target miscoverage $\alpha\in(0,1)$ define the integer cap
\begin{equation}
\tau_\alpha(r)\;=\;\big\lfloor \alpha\,r \big\rfloor .
\end{equation}
The window-level, class-conditional \emph{success} probability of meeting the $\alpha$-budget in one window is
\begin{equation}
p_{\mathrm{good}}(\alpha)
\;=\;
\Pr\!\big(e_j \le \tau_\alpha(m_j)\big)
\;=\;
\sum_{r=0}^{m}\Pr(m_j=r)\;\Pr\!\big(e_j \le \tau_\alpha(r)\,\big|\,m_j=r\big),
\end{equation}
which Algorithm~\ref{algo:ssbc} evaluates (in place of a Beta tail) on the discrete grid $\{u/(n_j+1)\}$, accepting the largest $\alpha_{\mathrm{adj}}<\alpha_{\text{target}}$ with $p_{\mathrm{good}}(\alpha_{\text{target}})\ge 1-\delta$. We do not marginalize over $m_j$ first: the event $e_j \le \tau_\alpha(m_j)$ couples $e_j$ and $m_j$, and using only the marginal law of $e_j$ miscomputes this probability, especially when $m_j$ can be small.

For comparison, if the goal is the canonical per-item Mondrian guarantee (conditional on class $j$), prevalence does not enter the target; one simply applies Algorithm~\ref{algo:ssbc} with $n$ replaced by $n_j$ (and, if desired, its finite-$m$ Beta--Binomial analogue), thereby controlling $\Pr(\text{coverage}\ge 1-\alpha_{\text{target}})$ over calibration randomness within class $j$. 

\section*{Appendix:Molecular Solubility Network}

The molecular solubility prediction model employs a compact transformer-based architecture specifically designed for quantile regression on molecular graph representations. The network processes molecules as sequences of atoms with their chemical properties.
\paragraph{Molecular Feature Extraction:} Molecules are converted from SMILES strings to atom-level features using RDKit. Each atom is represented by four key properties: atomic number, degree (number of bonds), formal charge, and hybridization state. These features are embedded using separate embedding layers of dimension 8 and concatenated before projection to the hidden dimension.
\paragraph{Transformer Encoder}: The embedded atom sequences are processed through an 8-layer transformer encoder with 2 attention heads and feed-forward dimensions of 16. Each layer uses pre-normalization (LayerNorm before attention/FFN) and GELU activation for improved stability. Positional encodings are added to capture sequential information in the molecular representation.
\paragraph{Global Molecular Representation}: The transformer output uses a hybrid pooling strategy combining attention-weighted pooling with mean pooling, controlled by the parameter $\lambda_\mathrm{attention} = 0.5$. This allows the model to balance between learned importance weighting and uniform aggregation across atoms.
\paragraph{Quantile-Constrained Prediction}: Rather than independent quantile heads, the architecture uses a constrained approach with three components: a median predictor and two delta predictors for deviations. The median head predicts the 50th percentile directly, while delta heads predict positive offsets using softplus activation. This ensures quantile ordering: $q_{25} = \mathrm{median} - \Delta_{\mathrm{low}}$, $q_{75} = \mathrm{median} + \Delta_\mathrm{high}$, guaranteeing $q_{25} \le q_{50} \le q_{75}$.
\paragraph{Training Protocol}: The model uses randomized pinball loss, randomly selecting one quantile per batch for gradient computation. Training employs Adam optimization with learning rate 0.01, gradient clipping (max norm = 1.0), and \emph{ReduceLROnPlateau} scheduling. The compact architecture (8-dimensional hidden state) enables efficient training while maintaining predictive accuracy for molecular solubility estimation.

More sophisticated molecular representations could be constructed using pre-trained atom embeddings from models like ChemBERTa or OpenMolecules, rather than the simple hybridization-based features employed here. The goal of this compact network architecture was to demonstrate the practical utility of SSBC in a quantile regression setting while maintaining computational efficiency for the conformal prediction experiments.

\end{document}